Reconstructing MODIS Normalized Difference Snow Index Product on Greenland Ice Sheet Using Spatiotemporal Extreme Gradient Boosting Model

Fan Ye, Qing Cheng, Weifeng Hao, Dayu Yu

# Reconstructing MODIS Normalized Difference Snow Index Product on Greenland Ice Sheet Using Spatiotemporal Extreme Gradient Boosting Model

Fan Ye [1], Qing Cheng[1,*], Weifeng Hao[2], Dayu Yu[3]

[1] School of Computer Science, China University of Geosciences, Wuhan 430074, China

[2] Chinese Antarctic Center of Surveying and Mapping, Wuhan University, Wuhan 430079, China

[3] School of Remote Sensing and Information Engineering, Wuhan University, Wuhan 430079, China

**ABSTRACT**

The spatiotemporally continuous data of normalized difference snow index (NDSI) are key to understanding the mechanisms of snow occurrence and development as well as the patterns of snow distribution changes. However, the presence of clouds, particularly prevalent in polar regions such as the Greenland Ice Sheet (GrIS), introduces a significant number of missing pixels in the MODIS NDSI daily data. To address this issue, this study proposes the utilization of a spatiotemporal extreme gradient boosting (STXGBoost) model generate a comprehensive NDSI dataset. In the proposed model, various input variables are carefully selected, encompassing terrain features, geometry-related parameters, and surface property variables. Moreover, the model incorporates spatiotemporal variation information, enhancing its capacity for reconstructing the NDSI dataset. Verification results demonstrate the efficacy of the STXGBoost model, with a coefficient of determination of 0.962, root mean square error of 0.030, mean absolute error of 0.011, and negligible bias (0.0001). Furthermore, simulation comparisons involving missing data and cross-validation with Landsat NDSI data illustrate the model's capability to accurately reconstruct the spatial distribution of NDSI data. Notably, the proposed model surpasses the performance of traditional machine learning models, showcasing superior NDSI predictive capabilities.

This study highlights the potential of leveraging auxiliary data to reconstruct NDSI in GrIS, with implications for broader applications in other regions. The findings offer valuable insights for the reconstruction of NDSI remote sensing data, contributing to the further understanding of spatiotemporal dynamics in snow-covered regions.

**1. Introduction**

Snow cover is a vital component of the global climate system, distributed widely in middle and high latitudes and alpine regions of the Northern Hemisphere (NH). Snow cover acts as an essential indicator of global climate change because of its high sensitivity and activity (Brown et al. 1996; Rupp et al. 2013). The surface area and duration of snow cover directly affect surface radiation, heat balance, and energy exchange of the ground-air system. According to the 2021 assessment report by the Intergovernmental Panel on Climate Change, the cryosphere is experiencing an accelerated shrinkage, and snow cover extent (SCE) in the NH is decreasing rapidly (Arias et al. 2021), which has a considerable impact on the energy balance and water cycle of the NH (Born et al. 2019). Therefore, obtaining spatial and temporal continuous long-term snow cover data is critical for regional and global climate prediction and water resource management.

Remote sensing technology provides high-quality and effective data for continuous monitoring of the spatial and temporal behavior of snow cover, and the commonly used snow cover remote sensing data sources include NOAA AVHRR (Naegeli et al. 2021), MODIS (Riggs and Hall, 2015), VIIRS (Hall et al. 2019), and other series of satellite data (as shown in Table 1). Among them, MODIS is widely used because of its multispectral band placement, radiometric resolution that reduces saturation over bright snow, high spatial and temporal resolution, and long recording time span (Li et al. 2019). The MODIS

Collection 5 provides snow cover area (SCA) and fractional snow cover (FSC) data, and the Collection 6 (C6) product, which was released in 2016, provides Normalized Difference Snow Index (NDSI) data. The NDSI is related to the presence of snow in pixels, thereby describing snow fraction more accurately than SCA and FSC do (Riggs et al. 2017; Hall et al. 2019). Compared to in-situ measurements and higher-resolution Landsat and Sentinel data, the data of MODIS C6 NDSI product has reliable accuracy under clear sky conditions (Crawford, 2015; Zhang et al. 2019; Aalstad et al. 2020), and has been successfully applied in snow mapping and hydrological modeling (Dong, 2018; Tong et al. 2020; Tang et al. 2022; Luo et al. 2022; Deng et al. 2024). Unfortunately, NDSI products suffer from serious missing data problems caused by cloud contamination. Thus, developing a method to reconstruct spatiotemporally continuous NDSI based on the MODIS C6 products is crucial.

**Table 1** Information on widely used NDSI remote sensing products.

| Product | Temporal resolution | Spatial resolution | Time range |
| --- | --- | --- | --- |
| AVHRR | 1 day | 25 km | 1981 - now |
| VEGETATION | 10 day | 1 km | 1999 - 2012 |
| Landsat | 16 day | 30 m | 1972 - now |
| Sentinel | 5 day | 10 m | 2015 - now |
| VIIRS | <1 day | 750 m | 2012 - now |
| MODIS | 1 day | 500 m | 2000 - now |

Several algorithms have been developed for cloud removal and NDSI information reconstruction to solve the cloud contamination problem in MODIS C6 NDSI products. Some studies have addressed the issue of missing pixels in MODIS NDSI by selecting similar pixels to fill the gaps. For instance, Hou et al. (2019), Li et al. (2020), and Chen et al. (2020) proposed the use of nonlocal spatiotemporal filtering, conditional probability, and spatiotemporal adaptive gap filling to reconstruct data. Jing et al. (2022) developed a two-stage fusion technology using Gaussian kernel functions to fill in cloud pixels and generate a cloud-free NDSI product for China from 2001 to 2020. Deep learning networks were explored in snow cover product reconstruction, as in the works of Xing et al. (2022) and Hou et al. (2022), who respectively used the U-Net model and long short-term memory network to recover missing data from MODIS NDSI. However, relying solely on the geospatial–temporal correlation of NDSI can be inadequate in effectively addressing large-scale missing data issues over extended periods, particularly in polar regions where persistent cloud cover results in a significant amount of missing NDSI data that is difficult to reconstruct using existing methods.

In recent years, the use of multi-source ancillary data has been successful in the regression estimation of remotely sensed observations at regional and global scales. This approach has been applied successfully to reconstruct various environmental parameters such as PM2.5, surface temperature, chlorophyll concentration, and snow depth data (Wei et al. 2021; Zhao and Duan, 2020; Yang et al. 2020; Xiao et al. 2023; Wang et al. 2023). As for the snow cover related parameters, their spatial heterogeneity is intricately shaped by critical external environmental factors. Previous research has underscored the pivotal role of topographic characteristics, including elevation, slope, and aspect, in shaping the spatial heterogeneity of snowpack distribution (Revuelto et al., 2020). For instance, the High Mountain Asia region exhibits altitude-dependent variations in snow phenology, with higher-altitude mountains experiencing more snow-covered days, earlier snow onset, later snowmelt dates, and an extended snow cover duration compared to lower-altitude areas (Tang et al., 2022). Noteworthy studies by Koutantou et al. (2022) and Mazzotti et al. (2023) have observed significant dynamics in snow distribution on forested slopes, with higher variability and complexity on slopes exposed to solar radiation. Albedo, varying among different land cover types, further contributes to the complexity of snow cover's spatiotemporal distribution (Sieber et al., 2019). Even within snow and ice-covered areas, variations in albedo exist, as revealed during seasonal snow melting, exposing bare ice characterized by different albedo values (Alexander et al., 2014; Wehrlé et al., 2021). The interplay of these physical properties, redistribution dynamics, and external factors underscores the comprehensive considerations essential for reconstructing remote sensing data related to snow cover. Remote sensing estimation models based on multiple independent variables have accurately captured complex nonlinear relationships and offered the advantage of fast computation. However, the potential utility of this approach in the spatial and temporal prediction of NDSI has not been thoroughly investigated, and its efficacy remains unclear.

  Machine learning techniques have been particularly effective in unraveling complex nonlinear relationships among multiple variables. The extreme gradient boosting (XGBoost) algorithm (Chen and He, 2015) stands out as a scalable tree boosting system that effectively addresses challenges encountered by traditional machine learning algorithms. These challenges include high sensitivity to sample data, computational complexity, and the risk of model overfitting (Lin et al. 2023). XGBoost is widely embraced by data scientists for its robust performance (Li et al. 2020). In contrast to kernel-based methods like support vector machines (SVM), XGBoost offers a distinctive advantage by eliminating the need to map data into a high-dimensional space, thus reducing sensitivity to feature scaling. Furthermore, XGBoost's utilization of gradient boosting technology sets it apart from Random Forest (RF), as it iteratively trains weak learners, focusing on residuals to enhance prediction performance gradually. This iterative approach contrasts with RF, which relies on the ensemble of multiple decision trees and their collective votes (Jang et al., 2022). Particularly notable is XGBoost's efficiency in handling large-scale datasets, addressing challenges of computing speed and accuracy, ultimately resulting in reduced training and prediction times (Pilaš et al., 2020). In this context, based on the XGBoost algorithm, this study proposes an advanced spatiotemporal model to reconstruct MODIS NDSI daily data for regions with severe spatiotemporal gaps. Independent variables, including topographic, geometric correlation, and surface attribute variables, are incorporated into the model. Moreover, the spatiotemporal information is introduced to account for the heterogeneity of NDSI variability. The study aims to (1) develop a spatiotemporal XGBoost model to predict NDSI data, (2) evaluate the feasibility of using different auxiliary data to estimate the NDSI, and (3) assess the effectiveness of the proposed model through sample-based validation, simulated missing comparisons, and cross-validation with Landsat NDSI data in the Greenland Ice Sheet (GrIS).

**2. Materials and methods**

*2.1 Study area*

GrIS, is the largest ice sheet in the NH, covering an area of approximately 1.8 million km$^2$ and reaching a thickness of up to 3 km, with a water storage capacity of approximately 2.9 million km$^3$. The melting of this ice sheet could lead to a rise in the global sea level of approximately 7.2 m (Aschwanden et al. 2019). One of the most distinguishing features of the snow cover in GrIS is its extensive coverage over the surface, in contrast to other regions. This snow accumulation and melting pattern is heavily influenced by factors such as North Atlantic Oscillation and ocean heat transport. In recent decades, the air and ocean temperatures around GrIS have risen significantly, leading to reduced summer cloud cover (Hofer et al. 2017). These changes have resulted in widespread variations in surface elevation, including the formation of glacier lakes, glacier retreats, sub-ice melting, and an increase in ice streams, particularly near the margins of GrIS (Hugonnet et al. 2021). The complex geographical conditions and fragile ecological environment pose challenges in observing and simulating the snow cover in the region. Moreover, the interaction between surface eco-hydrological processes and regional climate is highly complicated.

The study area manifests discernible changes in the temporal dynamics of snow cover, particularly in the context of melting patterns. Recent investigations reveal that approximately 61% of designated melting areas within GrIS have experienced a notable extension in their melting duration. Conversely, around 38% of these areas have witnessed a reduction in melting duration, reflecting an overarching trend toward increased melting duration across the majority of the GrIS (Liang et al., 2019). This temporal shift is marked by earlier onset times of melt events and delayed dates of freeze initiation, collectively contributing to a prolonged melt season. Satellite observations corroborate these findings, illustrating distinct seasonal patterns in GrIS surface melting. Notably, the onset of melting is observed to commence in early May, reaching its peak around mid-July, and persisting until the conclusion of September (Zheng et al., 2022). Spatially, the study area exhibits noteworthy variations in the elevation of the end-season snowline, emphasizing regional heterogeneity in snow cover dynamics. Analyses reveal a consistent trend of increasing end-season snowline elevation with decreasing latitude, with values ranging from 980±67 meters in the northern regions to 1520±113 meters in the southwest (Ryan et al., 2019). In the specific case of GrIS, the NDSI method has been applied to identify various snow-related parameters, such as snowlines (Banwell et al. 2012), SCE (Cui et al. 2023), snow-free period length (Thompson et al. 2018), and glacier margins (Seale et al. 2011; Mallalieu et al. 2021). Additionally, a combination of NDSI and normalized water index has proven effective in accurately extracting GrIS glacial lakes (Shugar et al. 2020; Jiang et al. 2022). The hydrological balance of the cryosphere is greatly affected by the spatiotemporal variability of snow precipitation, seasonal SCE, and snow cover duration, which directly affects the variability of river runoff and glacier surface mass balance conditions (Malmros et al. 2018). Therefore, the acquisition of accurate NDSI data through remote sensing techniques is critical to analyze the distribution and changes in snow cover. This information is essential to an improved understanding of regional and global climate predictions and water resource management.

*2.2 Dataset*

In the course of this study, various auxiliary data were gathered, encompassing topographic, geometry-related, and surface property variables, all of which may influence the NDSI, incorporating topographic, geometry-related, and surface property variables. The topographic variables included elevation, aspect, and slope. The elevation data utilized in this study was derived from the GrIS digital elevation model (DEM), with a spatial resolution of 500 m, generated by Fan et al. (2022) based on observations from ICESat-2. The study confirmed the exceptional accuracy and stability of this data across diverse terrain conditions. For our purposes, ICESat-2 DEM data were employed to further derive GrIS slope and aspect data.

Geometry-related variables consisted of the sun and sensor zenith angles, along with the sun and sensor azimuth angles. Additionally, recognizing the potential influence of the surface environment on snow cover, auxiliary information such as land cover type and surface albedo was incorporated. Specifically, we used a spatiotemporal seamless surface albedo dataset generated by coupling spatiotemporal and physics-informed models for the GrIS (Ye et al., 2023). This approach maximizes the utilization of spatiotemporal information while accounting for the physical impact of clouds on ice and snow albedo. The accuracy of the reconstructed albedo was assessed through comparison with in-situ albedo measurements, revealing high accuracy.

Table 2 presents a comprehensive summary of the information and data sources integral to this study. Notably, all variables mentioned underwent a resampling process to align with the resolution of the MOD10A1 NDSI data. This meticulous effort was undertaken to ensure consistency and compatibility in subsequent analyses. For the purpose of this study, seven distinct tiles were selected: H15V02, H16V00, H16V01, H16V02, H17V00, H17V01, and H17V02. These tiles represent specific spatial regions, contributing to the spatial diversity considered in our investigation. The strategic use of these tiles enhances the robustness and representativeness of our findings, allowing for a nuanced exploration of spatiotemporal dynamics in the context of snow cover and NDSI.

**Table 2** Information of auxiliary data.

| Variable | Data source | Spatial resolution | Temporal resolution |
|---|---|---|---|
| Elevation | | 500 m | / |
| Aspect | Greenland DEM | 500 m | / |
| Slope | | 500 m | / |
| Land cover type | MCD12Q1 | 500 m | year |
| Solar zenith angle |  | 1 km | day |
| Sensor zenith angle | MOD09GA | 1 km | day |
| Solar azimuth angle |  | 1 km | day |
| Sensor azimuth angle |  | 1 km | day |
| Surface albedo | MOD10A1 | 500 m | day |

*2.3 Spatiotemporal XGBoost model*

The XGBoost model is an enhanced version of the gradient boosting decision tree, which is a strong learner combined with multiple weak learners. This model has the advantages of high accuracy, less overfitting, and scalability. The XGBoost model integrates multiple regression tree (CART) to compensate for the inability of a single CART to meet the prediction accuracy, and the prediction result is equal to the sum of the scores of all CARTs (Wu et al. 2019). The model is expressed in the following equation:

$$y_i = \sum_{n=1}^{N} f_k(x_i), f_k \in F, \tag{1}$$

where $i$ denotes the number of samples, $x_i$ denotes the eigenvalues of the samples, $y_i$ denotes the predicted values, $k$ denotes the number of regression trees, and $F$ denotes the set of CART.

The XGBoost model reduces the complexity of the approach and the risk of overfitting by performing a second-order Taylor expansion on the loss function and adding the regularization term to the objective function. To further improve the accuracy of reconstructed NDSI data, we introduce spatiotemporal auxiliary information into the XGBoost model and establish the STXGBoost model (Fig. 1). The temporal auxiliary information includes day (expressed as the day of the year) and the NDSI information of adjacent days. The spatial auxiliary information includes the longitude and latitude and the NDSI information of the neighboring spaces. Thus, this regression problem can be formulated as follows:

$$\text{NDSI} = f_{STXGBoost}(\text{Day, Lat, Lon, DEM, Asp, Slo, LAC, SoA, SoZ, SeA, SeZ, Albedo, SN, TN}), \quad (2)$$

$$\mathbf{SN} = \left(\sum_{M=m-1, N=n-1}^{M=m+1, N=n+1} y_{M,N,t} - y_{m,n,t}\right)/num, \quad (3)$$

$$\mathbf{TN} = \left(\sum_{T=t-1}^{T=t+1} y_{m,n,T} - y_{m,n,t}\right)/num, \quad (4)$$

where Day denotes the day of the year; Lat and Lon denote latitude and longitude, respectively; DEM, Asp and Slo denotes elevation, aspect and slope, respectively; LAC denotes land cover type; SoA, SoZ, SeA, and SeZ denote solar zenith angle, solar azimuth angle, sensor zenith angle, and sensor azimuth angle, respectively; Albedo denotes surface albedo, $(m, n)$ denotes the $(m, n)$th pixel at the spatial location, $t$ denotes the ordinal number of the pixel with temporal order $t$, $num$ denotes the number of valid pixels, $y_{m,n,t}$ denotes the NDSI value with spatial location $(m, n)$ on day $t$, SN denotes the average value of the valid NDSI in the adjacent spaces, TN denotes the average value of the valid NDSI in the adjacent days, and NDSI represents the predicted NDSI value.

The XGBoost model automatically provides a relative importance index for each feature, and can obtain the ranking of the importance of all features, as shown in Fig. 2. The model prioritizes SN as the most influential contributor to NDSI estimation. This strategic choice aligns with geographical principles, as the NDSI information from neighboring spatial locations stands out as the primary reference for accurate predictions. Following this, the model incorporates TN and Day variable to capture the temporal dynamics inherent in NDSI patterns. In further enhancing prediction reliability, the inclusion of the albedo factor is motivated by the physical properties of snow, characterized by high reflectivity. This addition contributes to refining NDSI predictions by leveraging the inherent reflective properties of snow surfaces. Subsequently, the model incorporates geographical location information, surface type, aspect, slope, and elevation, acknowledging the intricate spatial complexity and heterogeneity of pixels. These variables provide detailed insights into surface characteristics, enriching the predictive capacity of the model. Finally, obstruction in optical satellite observations because of topography may affect the NDSI estimation.

Geometric correlation variables can reflect the variation of surface reflectance to a certain extent, and the geometric correlation variables are relatively less useful for predicting NDSI from the results.

The STXGBoost model can reconstruct spatiotemporally complete NDSI data, but it may contain some noise. One way to address this issue is to apply a Savitzky–Golay (SG) filter to the time series data. The benefit of using SG filtering is noise removal while the shape and width of the signal are preserved. Therefore, in this study, we applied SG filtering as a post-processing step to refine the NDSI time series data.

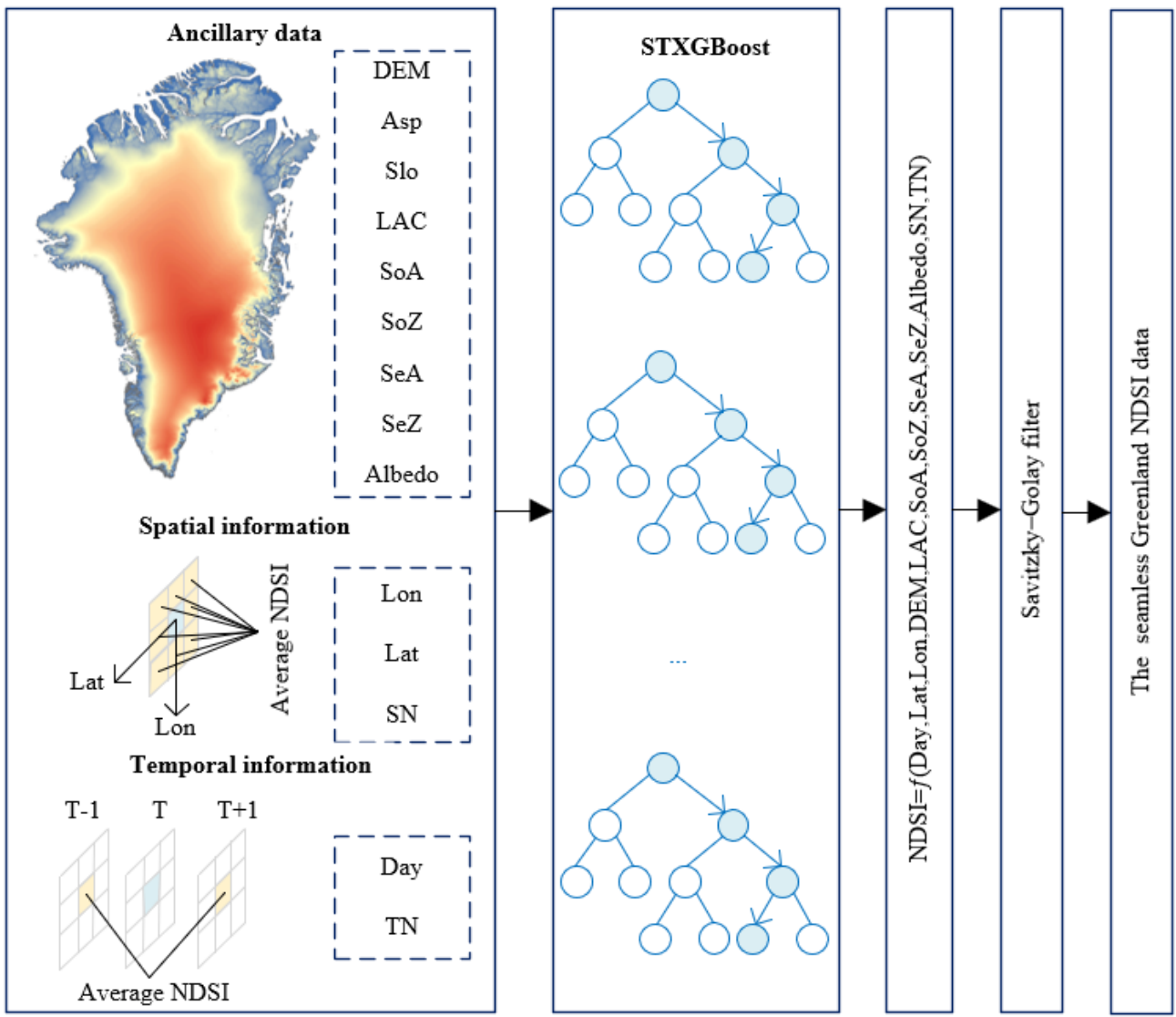

**Fig. 1.** Flow chart of proposed NDSI reconstruction model.

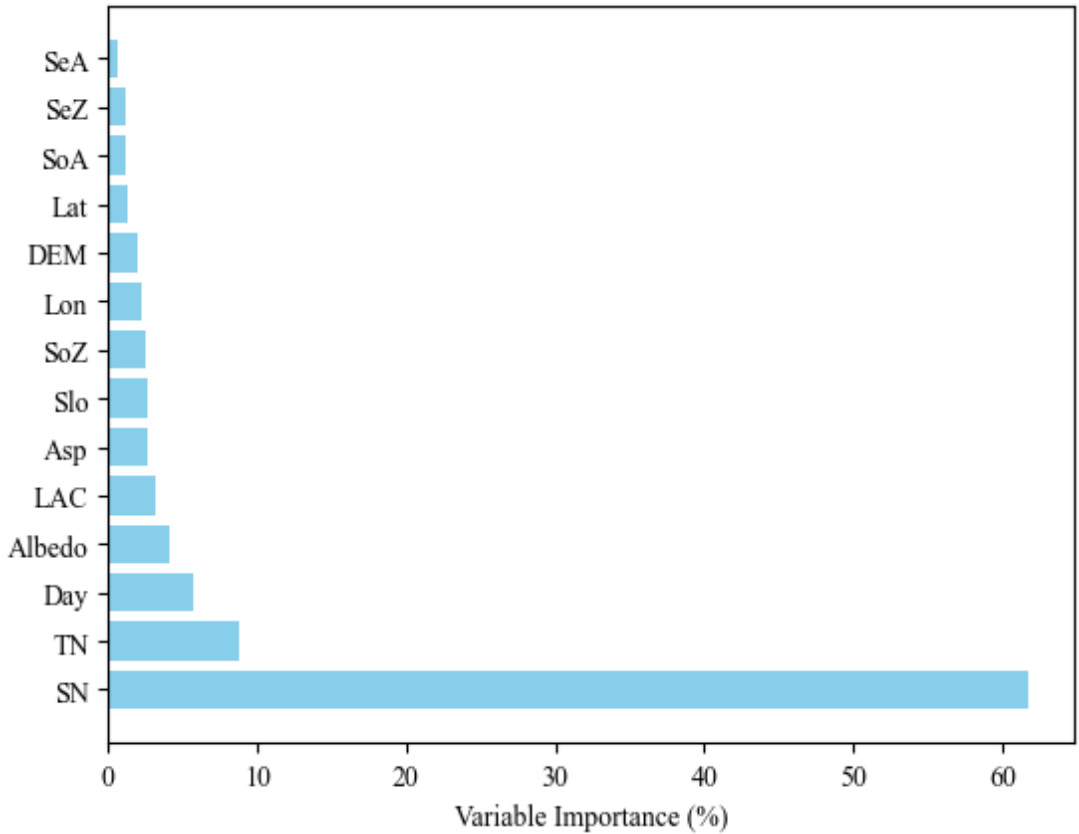

**Fig. 2.** Ranking of variable importance.

*2.4 Validation methodology*

In the GrIS region, snow depth data are provided by automatic weather stations. However, the zero point on the snow height axis of the automatic weather stations is arbitrary and set to zero in the first measurement at the station, measuring the height of the surface rather than the true snow depth. This limitation renders the in-situ measurement unsuitable for verifying the accuracy of the reconstructed data. To address this issue, we employed three evaluation methods: sample-based validation, missing data simulation comparison, and cross-validation with Landsat NDSI data. These methods were used to quantitatively evaluate the performance of the proposed NDSI missing pixel reconstruction model. Additionally, we used four evaluation metrics, including coefficient of determination ($R^2$), root-mean-square error (RMSE), mean absolute error (MAE), and mean difference error (Bias). The evaluation metrics are defined as follows:

$$R^2 = 1 - \frac{\sum_{i=1}^{n}(A_{rec}^i - \overline{A_{rec}})^2}{\sum_{i=1}^{n}(A_t^i - \overline{A_t})^2}, \tag{5}$$

$$RMSE = \sqrt{\frac{\sum_{i=1}^{n}(A_{rec}^i - A_t^i)^2}{n}}, \tag{6}$$

$$MAE = \frac{1}{n}\sum_{i=1}^{n}|A_{rec}^i - A_t^i|, \tag{7}$$

$$Bias = \frac{1}{n}\sum_{i=1}^{n}(A_{rec}^i - A_t^i), \tag{8}$$

where $n$ is the number of missing pixels; and $A_{rec}^i$ and $A_t^i$ are the $i$th reconstructed value and reference truth value, respectively. The reference true value can be Landsat NDSI or MODIS NDSI in test data.

## 3. Results

### 3.1 Predictive power of STXGBoost model

For the evaluation conducted on GrIS data from day 60 to day 299 of the year 2020, the STXGBoost model demonstrated robust predictive performance across different tiles. Utilizing a training set comprising 70% of the data and allocating the remaining 30% for testing, the model exhibited noteworthy accuracy. Table 3 presents the performance metrics, showcasing R² values consistently above 0.9, RMSE ranging between 0.020 and 0.067, MAE between 0.008 and 0.036, and Bias between -0.0001 and 0.0002. These outcomes affirm the STXGBoost model's ability to accurately predict NDSI data across diverse regions within GrIS. In aggregate, the model achieved an overall R² of 0.962, RMSE of 0.030, MAE of 0.011, and Bias of 0.0001, underscoring its robust predictive prowess throughout GrIS.

Further assessments were conducted to ascertain the model's predictive performance across different days, as illustrated in Fig. 3 Despite slight fluctuations in the four evaluation metrics over time, the overall prediction performance remained excellent. R² values ranged from 0.914 to 0.988, RMSE from 0.012 to 0.051, MAE from 0.006 to 0.023, and Bias from -0.00003 to 0.0004, with average values of 0.961, 0.029, 0.011, and 0.0001, respectively. These findings affirm the STXGBoost model's ability to predict NDSI data with precision across varying spatial missing rates and distinct temporal intervals within GrIS.

**Table 3** Predictive power of STXGBoost model generating NDSI for different tiles.

| Tile | R² | RMSE | MAE | Bias |
| --- | --- | --- | --- | --- |
| H15V02 | 0.976 | 0.026 | 0.011 | 0.0002 |
| H16V00 | 0.951 | 0.067 | 0.036 | 0.0001 |
| H16V01 | 0.962 | 0.021 | 0.009 | 0.0001 |
| H16V02 | 0.977 | 0.020 | 0.008 | 0.0001 |
| H17V00 | 0.939 | 0.046 | 0.020 | -0.0001 |
| H17V01 | 0.953 | 0.033 | 0.011 | 0.0000 |
| H17V02 | 0.908 | 0.049 | 0.019 | 0.0003 |
| ALL | 0.962 | 0.030 | 0.011 | 0.0001 |

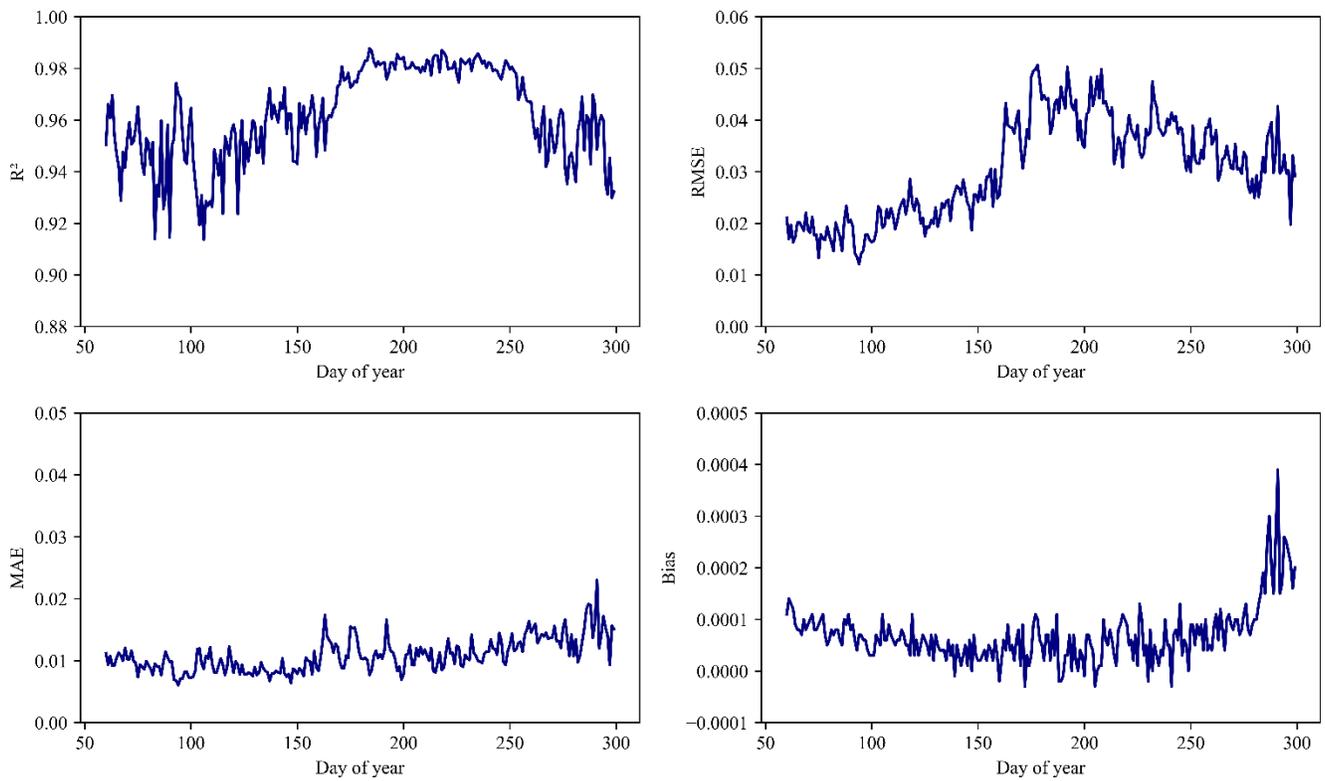

**Fig. 3.** Predictive power of STXGBoost model generating NDSI for different days.

*3.2 Validation based on simulated cloud mask*

To visually demonstrate the effectiveness of the STXGBoost model in reconstructing NDSI data, we simulated missing MODIS NDSI data by masking a certain percentage of data points to represent cloud coverage and then used the model to predict the missing values. Fig. 4 displays the data on MODIS NDSI, masked NDSI, reconstructed NDSI, and corresponding elevation. The figure shows that STXGBoost can effectively fill in the gaps in the NDSI images, and the reconstructed pixel distribution exhibits spatial continuity, accurately preserving the spatial distribution of snow in the real NDSI data. Comparison with the elevation data indicates that the reconstructed data can effectively recover the shape of the ridge, demonstrating the reasonable spatial distribution of the reconstructed data.

Furthermore, we conducted a quantitative evaluation of the STXGBoost reconstruction by comparing the reconstructed NDSI data with the MODIS NDSI data, as shown in Table 4. The results demonstrated that most of the $R^2$ values were around 0.9, and all of the RMSEs were less than 0.075, indicating the overall effectiveness of our approach in reconstructing the missing data. In the two images (3 and 4) of the edge region of GrIS, the reconstruction accuracy was slightly lower because of the lower altitude and rapid changes in the snow layer. However, the STXGBoost model still recovered the NDSI data stably, with the MAE maintained below 0.04.

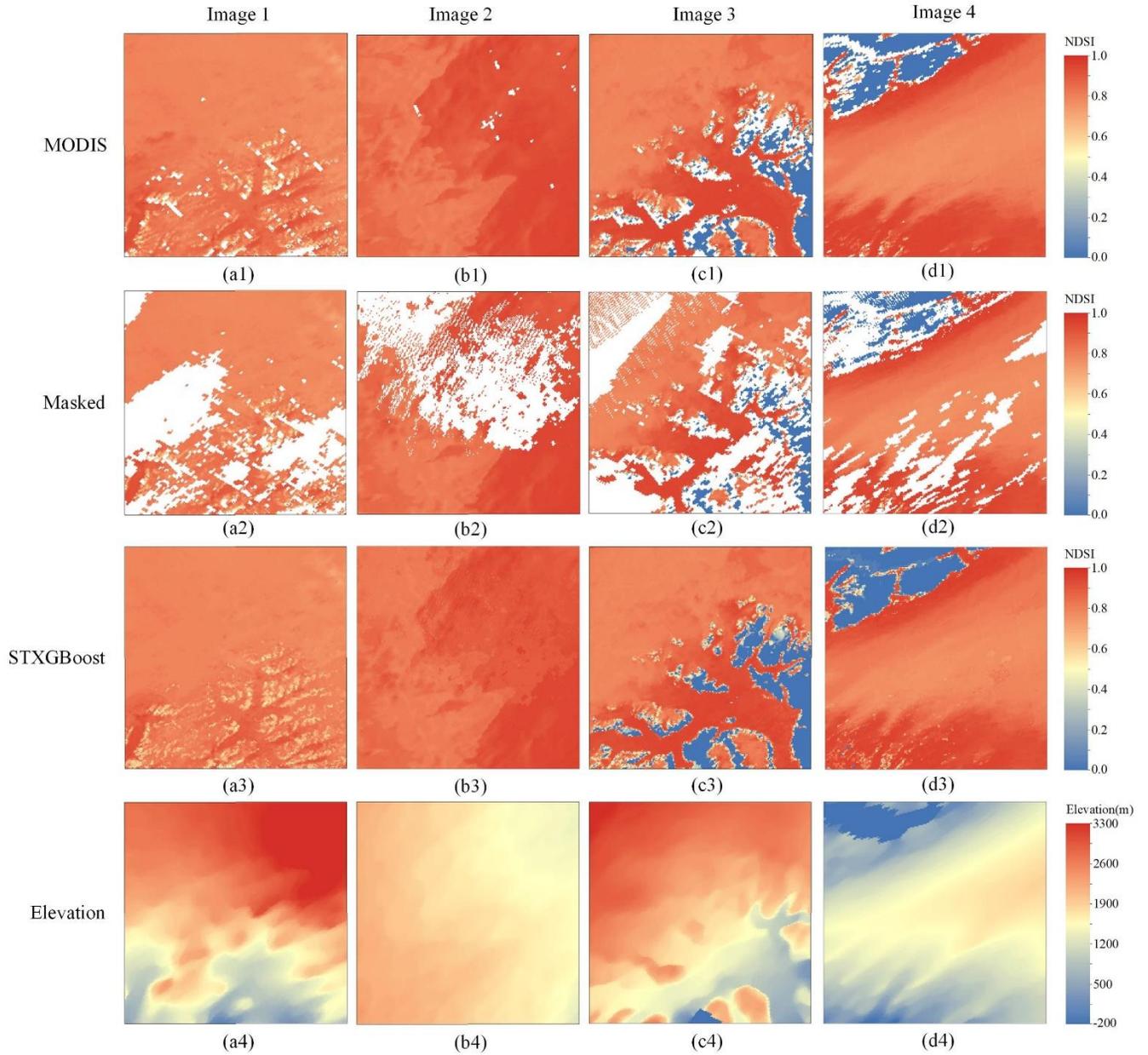

**Fig. 4.** Comparison of MODIS, masked, and reconstructed NDSI: (a1), (b1), (c1), and (d1) represent the MODIS data; (a2), (b2), (c2), and (d2) represent the simulated missing data; (a3), (b3), (c3), and (d3) represent the reconstructed data; (a4), (b4), (c4), and (d4) represent the corresponding elevation data.

**Table 4** Quantitative evaluation results of simulation experiments.

| Image | R² | RMSE | MAE | Bias | Mask Ratio (%) |
|---|---|---|---|---|---|
| 1 | 0.901 | 0.027 | 0.022 | 0.017 | 28.57 |
| 2 | 0.527 | 0.049 | 0.030 | 0.021 | 32.78 |
| 3 | 0.973 | 0.060 | 0.023 | 0.006 | 38.40 |
| 4 | 0.967 | 0.072 | 0.031 | 0.020 | 24.87 |

*3.3 Validation based on Landsat NDSI maps*

The higher spatial resolution Landsat NDSI data can be used to assess the quality of the reconstructed data. Since the Landsat 8 dataset does not completely cover GrIS, which prevents a comparison of the different parts of the entire study area, the areas with relatively significant changes in NDSI data were selected for comparison. To ensure consistency in spatial scale, we resampled the Landsat NDSI data to match the MODIS resolution. Fig. 5 shows the MODIS NDSI clear sky data (a1 and a2), reconstructed NDSI data (b1 and b2), and Landsat data (c1 and c2), respectively. The inconsistent wavelength ranges of the bands used by MODIS and Landsat to invert the NDSI data make the values of these two NDSI datasets different. As the figure shows, the spatial distribution of the reconstructed NDSI data and Landsat NDSI data in these two regions show a high degree of consistency. The reconstructed images accurately identify the NDSI dividing lines, indicating that the reconstructed data are satisfactory.

Table 5 shows the results of the quantitative evaluation of the MODIS and reconstructed data with Landsat data. The results show that compared with MODIS NDSI, the reconstructed data $R^2$ of areas 1 and 2 decreased by 0.048 and 0.077, respectively, and the RMSE increased by 0.010 and 0.025, respectively. The slight decrease in accuracy is reasonable because the reconstructed regions inherit the errors of MODIS clear-sky areas. Overall, the proposed model can reconstruct high-quality NDSI information.

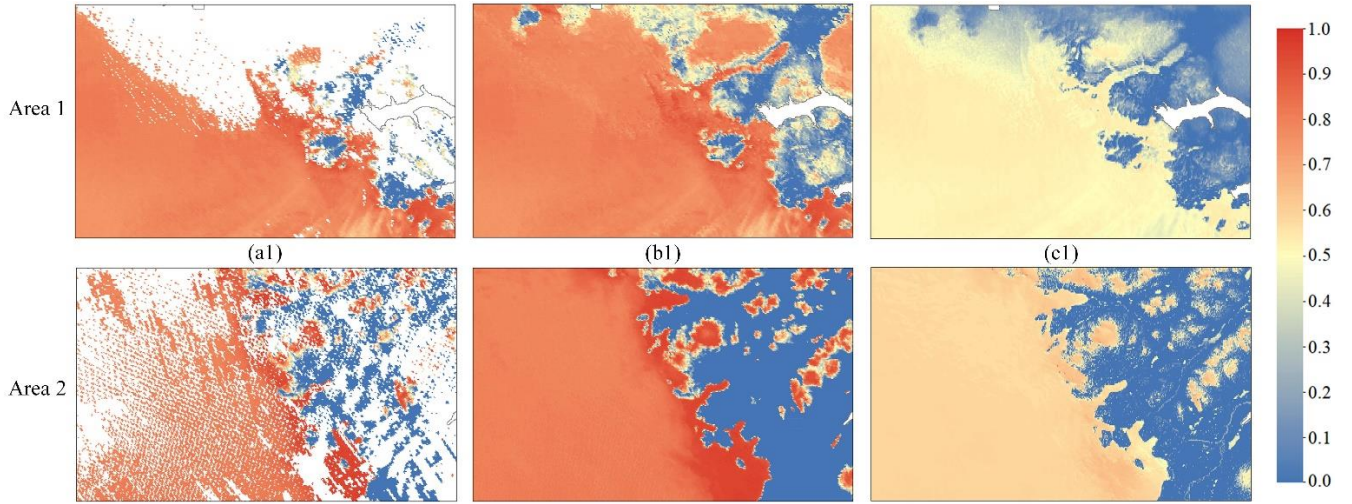

**Fig. 5.** Comparison of MODIS, reconstructed, and Landsat NDSI: (a1) and (a2) represent the MODIS NDSI data, (b1) and (b2) represent the reconstructed NDSI data, and (c1) and (c2) represent the Landsat NDSI data.

**Table 5** Quantitative evaluation results compared with Landsat data.

| Area | Data | $R^2$ | RMSE | MAE | Bias |
|---|---|---|---|---|---|
| 1 | MODIS NDSI | 0.817 | 0.247 | 0.230 | 0.226 |
|   | Reconstructed NDSI | 0.769 | 0.257 | 0.228 | 0.223 |
| 2 | MODIS NDSI | 0.738 | 0.266 | 0.256 | 0.252 |
|   | Reconstructed NDSI | 0.661 | 0.291 | 0.260 | 0.257 |

*3.4 Comparison with other models*

To assess the efficacy of the STXGBoost model, we conducted a comparative analysis with several widely employed regression models, namely Multiple Linear Regression (MLR), RF, SVM, and XGBoost. Additionally, we introduced two supplementary models, SXGBoost and TXGBoost, by integrating spatial and temporal auxiliary data into the XGBoost framework, respectively. This was undertaken to ascertain the influence of spatiotemporal information on the NDSI reconstruction process. The MLR, RF, SVM, XGBoost, SXGBoost and TXGBoost model used here is expressed as:

$$\text{NDSI} = a_1\text{DEM} + a_2\text{Asp} + a_3\text{Slo} + a_4\text{LAC} + a_5\text{SoA} + a_6\text{SoZ} + a_7\text{SeA} + a_8\text{SeZ} + a_9\text{Albedo} + b, \quad (9)$$

$$\text{NDSI} = f_{RF}(\text{DEM}, \text{Asp}, \text{Slo}, \text{LAC}, \text{SoA}, \text{SoZ}, \text{SeA}, \text{SeZ}, \text{Albedo}), \quad (10)$$

$$\text{NDSI} = f_{SVM}(\text{DEM, Asp, Slo, LAC, SoA, SoZ, SeA, SeZ, Albedo}), \quad (11)$$

$$\text{NDSI} = f_{XGBoost}(\text{DEM, Asp, Slo, LAC, SoA, SoZ, SeA, SeZ, Albedo}), \quad (12)$$

$$\text{NDSI} = f_{TXGBoost}(\text{Day, DEM, Asp, Slo, LAC, SoA, SoZ, SeA, SeZ, Albedo, TN}), \quad (13)$$

$$\text{NDSI} = f_{SXGBoost}(\text{Lat, Lon, DEM, Asp, Slo, LAC, SoA, SoZ, SeA, SeZ, Albedo, SN}), \quad (14)$$

where $a_1, \ldots, a_n$ are the regression coefficients and b is the intercept; Day denotes the day of the year; Lat and Lon denote latitude and longitude, respectively; DEM, Asp and Slo denotes elevation, aspect and slope, respectively; LAC denotes land cover type; SoA, SoZ, SeA, and SeZ denote solar zenith angle, solar azimuth angle, sensor zenith angle, and sensor azimuth angle, respectively; Albedo denotes surface albedo, SN denotes the average value of the valid NDSI in the adjacent spaces, TN denotes the average value of the valid NDSI in the adjacent days, and NDSI represents the predicted NDSI value.

Following the experimental approach outlined in Section 3.1, we normalized all input data and randomly partitioned the dataset, allocating 70% for training and reserving the remaining 30% for testing. Subsequently, these partitions were applied to train and assess the performance of the STXGBoost model alongside the aforementioned six comparison methods. Table 6 presents the predictive performance metrics for each model. The MLR model yielded the lowest R² and the highest RMSE, indicative of its inadequacy for reconstructing NDSI data. While the RF and SVM models exhibited enhanced performance, their accuracy slightly trailed that of the XGBoost series models. Notably, the SXGBoost and TXGBoost models demonstrated improved forecast accuracy by accounting for spatial and temporal variability, respectively. Comprehensive evaluation across metrics—R², RMSE, MAE, and Bias—revealed that the STXGBoost model outperformed all counterparts, yielding values of 0.962, 0.030, 0.011, and 0.0001, respectively. These outcomes underscore the significance of incorporating spatiotemporal information, affirming its substantial contribution to enhancing the model's predictive prowess and yielding more precise NDSI data.

**Table 6** Quantitative evaluation of prediction data from different models.

| Model | R² | RMSE | MAE | Bias |
|---|---|---|---|---|
| MLR | 0.311 | 0.117 | 0.077 | -0.0001 |
| RF | 0.894 | 0.044 | 0.027 | 0.0004 |
| SVM | 0.833 | 0.052 | 0.029 | 0.0002 |
| XGBoost | 0.915 | 0.043 | 0.018 | 0.0001 |
| TXGBoost | 0.916 | 0.041 | 0.016 | 0.0002 |
| SXGBoost | 0.950 | 0.032 | 0.012 | 0.0001 |
| STXGBoost | 0.962 | 0.030 | 0.011 | 0.0001 |

*3.5 Temporal extension of the model*

We have conducted additional experiments to extend the temporal validation of the STXGBoost model. In this extended validation, the STXGBoost model is applied to forecast NDSI for the year 2019, serving as an independent validation dataset separate from the original training and testing data of 2020. The results, presented in Fig. 6, depict the accuracy of the daily forecast NDSI data for 2019. The model's performance is evaluated using key metrics, including R², RMSE, MAE, and Bias. For the 2019 forecast, the STXGBoost model demonstrates strong applicability and robustness, with R² ranging from 0.887 to 0.986, RMSE ranging from 0.022 to 0.058, MAE ranging from 0.011 to 0.026, and Bias ranging from -0.00007 to 0.0002. The average values for these metrics are 0.953, 0.039, 0.017, and 0.00006, respectively. These results affirm the model's effectiveness in spatiotemporally reconstructing NDSI data beyond the original training period, providing additional evidence of its reliability and applicability across different temporal contexts.

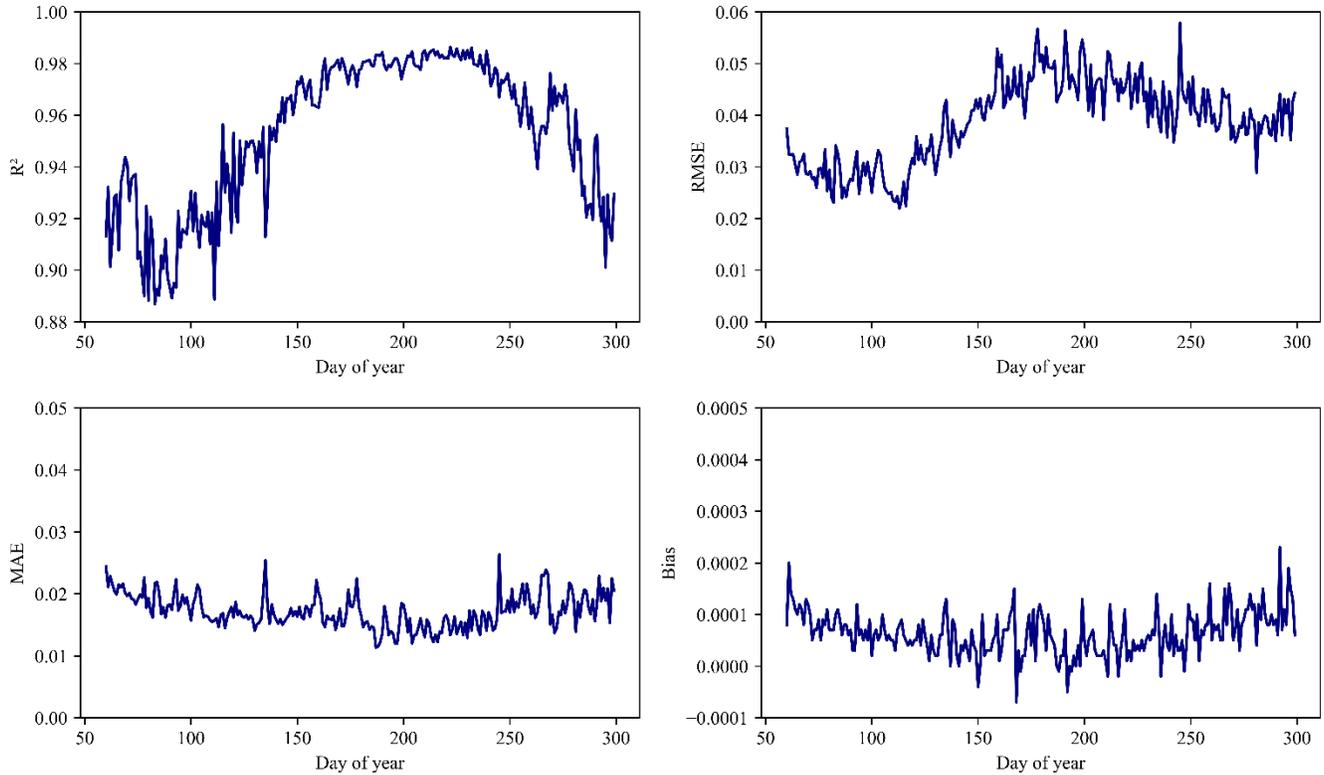

**Fig. 6.** Predictive power of STXGBoost model generating NDSI for 2019.

## 4. Discussion

### 4.1 Effect of STXGBoost model parameters

The STXGBoost model has three important parameters: learning_rate, max_depth, and n_estimators, which respectively control the learning rate, maximum depth, and number of decision trees in the model. To investigate the influence of these three parameters on prediction performance, we compared the predictive results under different parameter values, as shown in Fig. 7, where the range of max_depth is from 1 to 20, the range of learning_rate is from 0.01 to 0.2, and the range of n_estimators is from 50 to 450.

From the figure, we can observe that the predictive ability is weaker when the learning_rate is 0.01 and the n_estimators are less than 200. This condition may be due to the weak fitting ability of the model when the n_estimators are few, leading to the inability to effectively fit the complex relationships in the training data. While the predictive performance exhibits slight fluctuations with the change of these three parameters, the overall predictive results are relatively good, with most R² values above 0.8 and RMSE values mostly less than 0.075.

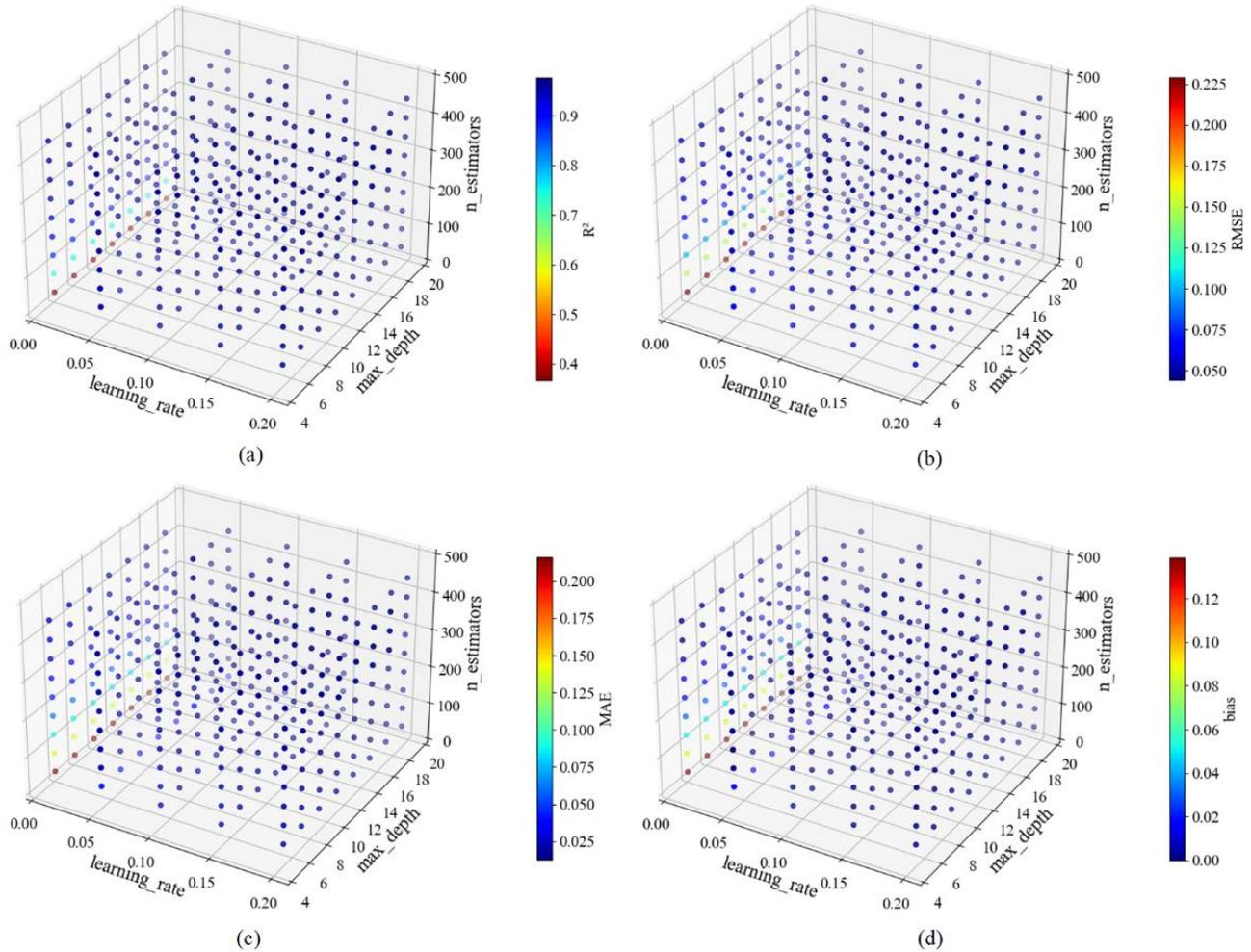

**Fig. 7.** Predictive performance with different parameter values.

*4.2 Effect of input variables on reconstruction performance*

In this study, various ancillary data were used to predict NDSI, including topographic, geometric correlation, and surface attribute variables. To analyze the effect of different combinations of independent variables on the models, we compared the predictive ability of NDSI using different combinations of auxiliary variables, and the results are presented in Table 7.

As depicted in the table, all the models demonstrated a predictive ability of NDSI with R² higher than 0.7, indicating the feasibility of reconstructing NDSI data with XGBoost. However, differences were observed in the prediction effects of different combinations of variables. Model 1, which considered only aspect, slope and DEM, demonstrated the worst reconstruction effect. By comparing model 1 with models 2, 3, and 4, we found that adding surface type, albedo, and geometric correlation data could further enhance the prediction accuracy, with R² increasing by 0.059, 0.147, and 0.063, and RMSE decreasing by 0.005, 0.007, and 0.005, respectively. These results indicate that adding albedo has a better prediction effect than adding surface type or geometric correlation. Comparing the results of Models 2, 4, and 5, we observed that considering both surface type and geometry-related variables improved the prediction performance. Furthermore, comparing the first five combinations shows that the models with albedo (models 3 and 6) have significantly better prediction results, with R² above 0.9 and RMSE below 0.04. Further comparison of all the different combinations of models together suggests that considering the adjacent spatial and temporal NDSI variations can enhance the prediction accuracy based on the consideration of multiple auxiliary data.

**Table 7** Effect of different auxiliary data selection on NDSI prediction results.

| Model | Data | R² | RMSE | MAE | Bias |
|---|---|---|---|---|---|
| 1 | Day+Lat+Lon+Asp+Slo+Dem | 0.763 | 0.046 | 0.019 | 0.0004 |
| 2 | Day+Lat+Lon+Asp+Slo+Dem+LAC | 0.822 | 0.041 | 0.017 | 0.0003 |
| 3 | Day+Lat+Lon+Asp+Slo+Dem+Albedo | 0.910 | 0.039 | 0.016 | 0.0001 |
| 4 | Day+Lat+Lon+Asp+Slo+Dem+SoZ+SeA+SeZ+SoA | 0.823 | 0.041 | 0.017 | 0.0001 |
| 5 | Day+Lat+Lon+Asp+Slo+Dem+LAC+SoZ+SeA+SeZ+SoA | 0.827 | 0.040 | 0.017 | 0.0002 |
| 6 | Day+Lat+Lon+Asp+Slo+Dem+LAC+SoZ+SeA+SeZ+SoA+Albedo | 0.926 | 0.038 | 0.015 | 0.0001 |
| 7 | Day+Lat+Lon+Asp+Slo+Dem+LAC+SoZ+SeA+SeZ+SoA+Albedo+SN+TN | 0.962 | 0.030 | 0.011 | 0.0001 |

*4.3 Role of training dataset size*

This section investigates the impact of varying proportions of training and test datasets on the predictive accuracy of the STXGBoost model reconstruction. The results are presented in Fig. 8. As anticipated, the training performance surpasses that of the test set, as the model parameters are intricately tuned to align with the characteristics of the training dataset. In instances where a smaller training dataset is employed, a marginally greater test error arises due to overfitting; however, the disparities in $R^2$, RMSE, MAE, and Bias between training and test accuracies are negligible, registering at 0.023, 0.016, 0.003, and 0.0005, respectively. Thus, overall, while the four evaluation indicators exhibit slight fluctuations with changing proportions of the test set, the predictive performance consistently maintains a satisfactory level. It is imperative to underscore that fluctuations in the evaluation indicators are expected as a consequence of adjusting the proportion of the test set. Nevertheless, the robustness of the STXGBoost model is evident, as it consistently delivers commendable predictive performance across varied training and test dataset proportions.

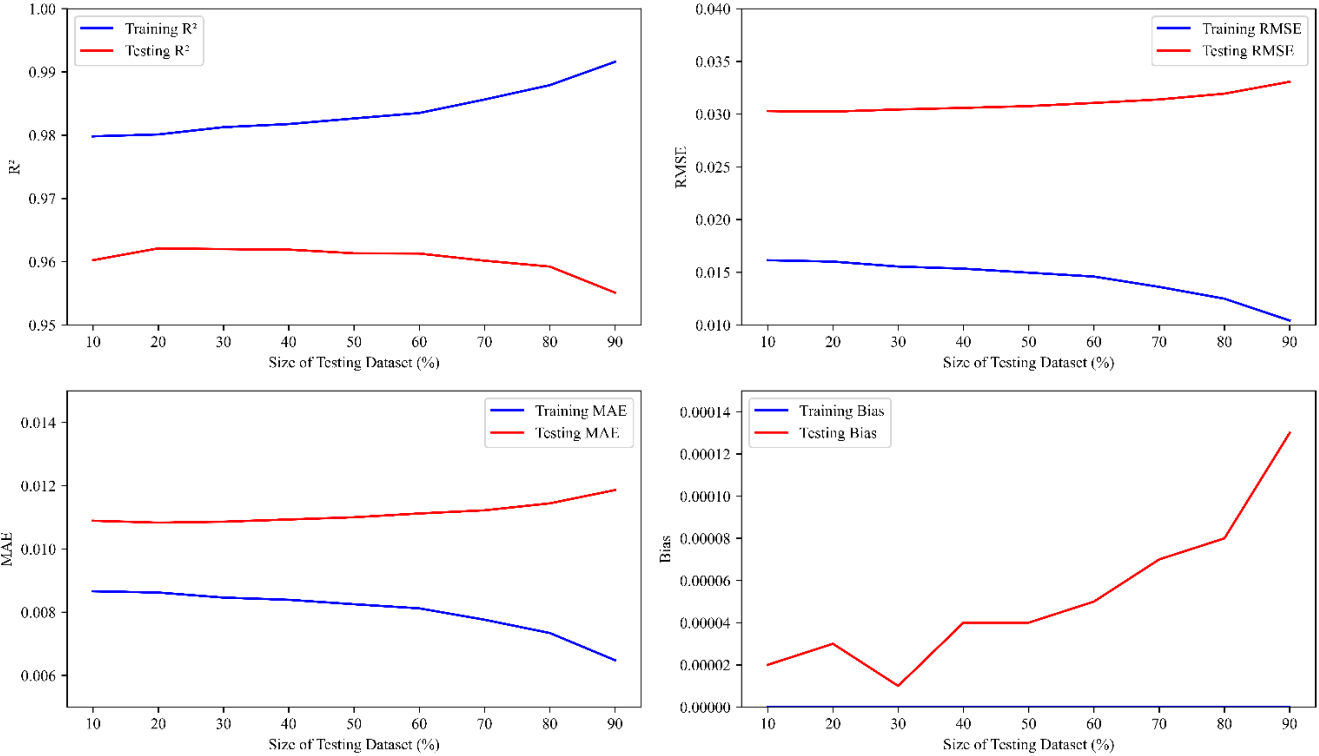

**Fig. 8.** Training and testing performance of STXGBoost.

*4.4 Sensitivity to temporal and spatial window lengths of SN and TN in the STXGBoost model*

The performance of the STXGBoost model is influenced by the size of the spatial window (SW) and temporal window (TW) associated with the input data SN and TN. In our investigation, we systematically explored the impact of varying SW and TW on the model's prediction efficacy, as depicted in Fig. 9. The results indicate an overall optimal performance within the ranges of SW from 1 to 10 and TW from 1 to 10. Notably, when SW is confined to the range of 1 to 2, minimal disparities are observed in the values of the four evaluation indicators—$R^2$, RMSE, MAE, and Bias. As TW increases, a subtle decrease in $R^2$ is noted, accompanied by slight elevations in RMSE and MAE. Conversely, when SW extends from 5 to 10, a more pronounced decline in $R^2$, coupled with noticeable increases in RMSE, MAE, and Bias, is evident. Consequently, our findings suggest that the most suitable values for SW and TW in the context of this study area are within the range of 1 to 2.

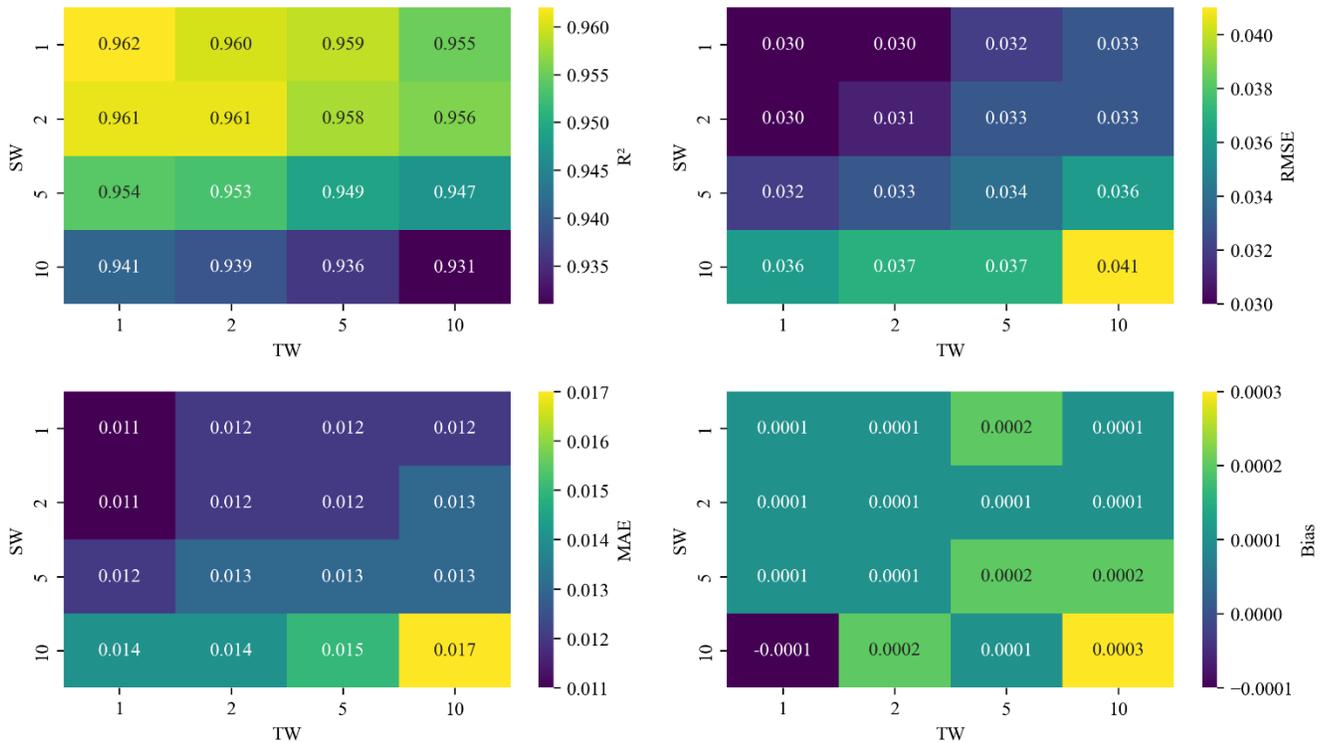

**Fig. 9.** The impact of SW and TW on the STXGBoost model.

*4.5 Scalability and applicability of the STXGBoost model to other regions*

The STXGBoost model proposed in this study demonstrates effective reconstruction of the spatiotemporal continuity of daily MODIS NDSI products for the GrIS. While our focus has been on GrIS, the model's scalability and applicability to other regions with similar data gaps or environmental conditions present promising avenues for future research. The STXGBoost model's adaptability to other regions, such as mountainous areas, forested regions, and urban environments, depends on the careful selection of auxiliary data tailored to specific regional characteristics. For instance, in forested areas, the dense vegetation canopy often obstructs snow signals, leading to inaccuracies in NDSI measurements. Integrating additional remote sensing data, such as Synthetic Aperture Radar (SAR), which can penetrate vegetation, may significantly enhance snow detection and reconstruction accuracy in these areas. Similarly, in mountainous regions where topography and rapid changes in snow cover are critical factors, incorporating high-resolution terrain data and snow depth measurements could improve model performance. Therefore, the potential application of the STXGBoost model to other regions (including vegetated areas) requires the integration of multi-source remote sensing data and other auxiliary variables to enhance the performance of the model in various geographical environments and make it applicable to a wider range of settings.

We recognize that any model-based approach may introduce errors. These biases can stem from various sources, including the quality of input data, specific regional characteristics, and the model's inherent assumptions. Future research can improve the reliability of reconstructed NDSI through uncertainty quantification, sensitivity analysis, and similar techniques. Additionally, exploring the impact of different parameter settings (e.g., learning_rate, max_depth, and n_estimators) on model performance is essential for optimizing the STXGBoost model for various scenarios. Systematic studies of these parameters through techniques such as grid search or random search can identify the optimal configurations that maximize model performance, thereby addressing potential biases and uncertainties in the reconstructed NDSI dataset.

## 5. Conclusion

Optical satellites are a valuable resource for monitoring the spatiotemporal variability of snow. However, their usefulness is limited because of cloud cover occlusion that results in many missing pieces of data. To overcome this limitation, this study proposed the STXGBoost model to reconstruct the spatiotemporal continuity of daily MODIS NDSI products. Our evaluation of the accuracy of the model included sample-based validation, simulation-based missing validation, and cross-validation with Landsat NDSI, all of which demonstrated the satisfactory reconstruction capability of the model. By comparing the effects of various input data on the prediction results, we found that the choice of auxiliary data had a significant effect on the prediction performance of the model, with spatial information providing the most substantial boost. Furthermore, compared to classical regression models, the XGBoost series demonstrates superior prediction ability, and the STXGBoost model recovers missing NDSI information more robustly and effectively by incorporating spatiotemporal information.

Overall, this study highlighted the feasibility of using ancillary data and spatiotemporal information to estimate NDSI, providing new insights into the reconstruction of snow remote sensing data. Furthermore, the spatiotemporal NDSI reconstruction approach proposed in this study contributed to a better understanding of the spatial and temporal patterns of snow cover in GrIS, and its relationship with long-term climate change.

**Acknowledgment**

This work was supported by the National Natural Science Foundation of China (Grant No. 42171383).